\documentclass[11pt]{article}

\usepackage[a4paper,top=2.1cm,bottom=2.1cm,left=2.2cm,right=2.2cm]{geometry}
\usepackage[T1]{fontenc}
\usepackage[utf8]{inputenc}
\usepackage{newtxtext}
\usepackage[amsthm]{newtxmath}
\usepackage{amsmath}
\usepackage{microtype}
\usepackage{booktabs,tabularx,array}
\usepackage{graphicx}
\usepackage{xcolor}
\usepackage[
  backend=biber,
  style=nature,
  sorting=none,
  natbib=true,
  maxbibnames=5,
  minbibnames=1
]{biblatex}
\usepackage[hidelinks]{hyperref}
\usepackage{authblk}
\usepackage{placeins}

\title{%

\huge{Concordance Plausibility Check for
Intervention-Conditioned Human Digital Twins}

}

\author{%
\begin{minipage}{0.94\textwidth}

\centering
Saeed~Farzi\textsuperscript{1},
Christopher~Kipps\textsuperscript{2,3},
John~McBeth\textsuperscript{4,5},
Gabriela~Czanner\textsuperscript{1,5,7},
Sofia~Michopoulou\textsuperscript{2,3},
Benjamin~Macarthur\textsuperscript{6,7},
Sarah~Ennis\textsuperscript{8},
Paul~Elkington\textsuperscript{9,10,11},
Nicholas~C~Harvey\textsuperscript{11,12},
Rahman~Attar\textsuperscript{1,*}

\par\vspace{1em}

\begingroup
\footnotesize

\textsuperscript{1}
School of Electronics and Computer Science,
University of Southampton, Southampton, UK
\par\vspace{0.25em}

\textsuperscript{2}
Faculty of Medicine,
University of Southampton,
Southampton, UK
\par\vspace{0.25em}

\textsuperscript{3}
University Hospital Southampton NHS Foundation Trust,
Southampton, UK
\par\vspace{0.25em}

\textsuperscript{4}
Primary Care Research Centre, Faculty of Medicine, University of Southampton, Southampton, UK
\par\vspace{0.25em}

\textsuperscript{5}
Digital Health and Biomedical Engineering Group, Faculty of Engineering and Physical Sciences,
University of Southampton, Southampton, UK
\par\vspace{0.25em}

\textsuperscript{6}
Mathematical Sciences,
University of Southampton, Southampton, UK
\par\vspace{0.25em}

\textsuperscript{7}
Human Development and Health,
Faculty of Medicine,
University of Southampton, Southampton, UK
\par\vspace{0.25em}

\textsuperscript{8}
Human Genomic Informatics Group,
Faculty of Medicine,
University of Southampton, Southampton, UK
\par\vspace{0.25em}

\textsuperscript{9}
Institute for Life Sciences,
University of Southampton, Southampton, UK
\par\vspace{0.25em}

\textsuperscript{10}
Clinical and Experimental Sciences Academic Unit,
Faculty of Medicine,
University of Southampton,
Southampton, UK
\par\vspace{0.25em}

\textsuperscript{11}
NIHR Biomedical Research Centre: Southampton,
University of Southampton and University Hospital
Southampton NHS Foundation Trust,
Southampton, UK

\textsuperscript{12}
Lifecourse Epidemiology Centre,
University of Southampton,
Southampton, UK
\par\vspace{0.25em}

\par\vspace{0.7em}

\textsuperscript{*}
Corresponding author: Rahman Attar (email: r.attar@southampton.ac.uk)

\par
\endgroup

\end{minipage}%
}

\begin{document}

\date{} 
\maketitle

\begin{abstract}
Intervention-conditioned human digital twins predict treatment alternatives that cannot be observed in the same patient. Ordinary predictive accuracy therefore cannot establish counterfactual credibility. We propose the Concordance Plausibility Check (CPC), a claim-centered framework combining real-world evidence, qualified simulation and clinical knowledge. CPC matches evidence to the causal claim, audits contradictions and shared dependencies, and reports Pass, Fail or reason-coded Indeterminate.
\end{abstract}

\section{Introduction}
Intervention-conditioned human digital twins (IC-HDTs) are advancing toward treatment comparison, trial design and personalized decision support \citep{laubenbacher2024digitaltwins,akbarialiabad2025trials,ruhrberg2026roadmap}. Their central evaluation problem is unusual. The clinically important output is often what might happen under a treatment that the patient did not actually receive. For example, this may be the predicted disease trajectory for the same patient had an alternative treatment been given. Ordinary predictive accuracy can therefore show whether a twin reproduces observed outcomes, but it cannot by itself establish the credibility of the unobserved alternative.

We use \emph{digital twin} consistently with the National Academies' operational definition: a virtual representation dynamically updated from its physical counterpart, with predictive capability and a decision-support role \citep{nasem2024digitaltwins}. Here, \emph{intervention-conditioned} means that the twin is queried under a specified treatment or policy and generates the corresponding future trajectory. This does not automatically make the output causal. A causal interpretation additionally requires a well-defined intervention, defensible assumptions about treatment assignment and support, and a justified connection between the model output and the clinical outcome that would occur under that intervention \citep{hernan2020whatif,robins2000msm,pearl2009causality}.

The recent Perspective by Ruhrberg Estévez et al.~\cite{ruhrberg2026roadmap} outlines how causal inference and digital twins can support clinical trials. Building on that roadmap, we address the complementary evaluation problem: what evidence is sufficient to support the counterfactual claims generated by an intervention-conditioned twin when the relevant alternative outcomes are unobserved.

This problem is particularly important for in silico clinical trials, where digital twins may construct virtual treatment arms or compare alternative intervention strategies \citep{akbarialiabad2025trials,ruhrberg2026roadmap}. The virtual trial is built from intervention-specific trajectories, but the alternative trajectory for the same real patient is unavailable. \textbf{The validity of an in silico clinical trial therefore depends critically on the credibility of the counterfactual trajectories on which it is built.}

\begin{figure*}[t]
    \centering
    \includegraphics[width=\linewidth]{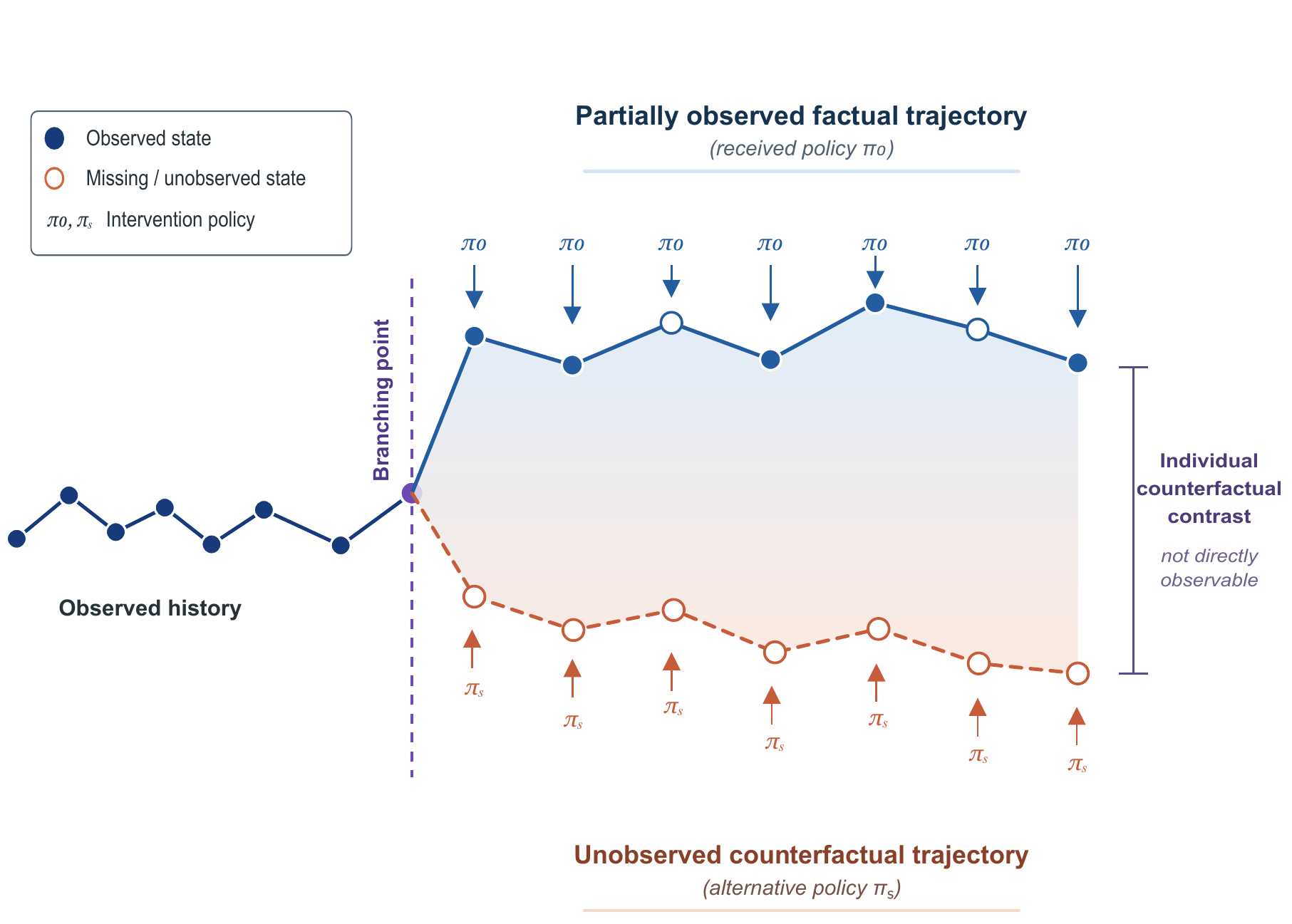}
    \caption{\textbf{The double-invisibility problem.}
    After a treatment decision, only one path is realized for a real patient, and that factual trajectory may itself be incompletely observed. The alternative intervention trajectory is not observed for the same patient. Evaluation must therefore distinguish evidence about the observed course from evidence that can constrain the unobserved alternative. Filled circles represent observed values, while open circles indicate missing values. The branching point marks where alternative interventions are applied}
    \label{fig:double_invisibility}
\end{figure*}

Evaluation is further complicated by a \emph{double-invisibility problem} depicted in Fig.~\ref{fig:double_invisibility}. Only one treatment path is realized for a patient \citep{holland1986causal}, and even that factual trajectory may be incompletely observed because of irregular visits, non-adherence, dropout, censoring, treatment switching, competing events, delayed entry, missing measurements or low-quality data. The evaluator therefore faces both an unobserved  counterfactual  world and an incomplete view of the factual world that occurred. Factual credibility is necessary, but it does not establish the validity of the counterfactual prediction.

Verification, validation and uncertainty quantification (VVUQ)
remain essential for credible clinical digital twins
\citep{sel2025vvuq,viceconti2020credibility,
pathmanathan2024credibility,asme2018vv40,fda2023credibility},
but they cannot by themselves prove that a counterfactual
trajectory is causally correct.
CPC builds on established methodological foundations:
V\&V40 links model credibility to context of use, and
ICH E9(R1) provides principles for clinical-trial estimand
specification \citep{asme2018vv40,ich2019e9r1}.
Cornish et al. contribute causal falsification, while
Laudy distinguishes marginal from cross-world validation
\citep{cornish2026causal,laudy2026dtcf}.
Vall\'ee's clinical-claim-based validation framework provides
an important conceptual foundation for matching validation
evidence to the intended clinical claim, complemented by
his work on decision-support credibility and causal
digital twins
\citep{vallee2026digital,vallee2026jamia,vallee2026causaltwins}. Complementary ethical assurance approaches, such as TEA-DT
\citep{burr2024trustworthy}, address broader trustworthiness
considerations, while CPC focuses specifically on evaluating
the plausibility of counterfactual claims under explicit
causal assumptions. 

Building on established approaches to model credibility, clinical-claim-based validation and causal falsification, CPC contributes an integrated, claim-specific evaluation framework for IC-HDTs. Its central contribution is to link explicitly registered counterfactual claims and their causal identification requirements to qualified evidence from observed data, simulation and clinical knowledge, while accounting for contradictions, uncertainty and shared evidence dependencies. These elements are combined through a non-compensatory decision rule that yields Pass, Fail or reason-coded Indeterminate outcomes. By distinguishing the evidential requirements of population-level, individual and structural counterfactual claims, CPC aims to prevent evidence supporting weaker claims from being used to justify stronger, unsupported conclusions. The resulting decisions are auditable, context-dependent and explicitly conditional on the available evidence and assumptions (Supplementary Table S1).

The organizing principle is simple: \textbf{evidence strength should match claim strength and causal structure}. CPC links a claim-to-assumption map to three streams with limited authority, a provenance audit and reason-coded non-compensatory decisions. Judgments remain evidence- and assumption-conditional.

\section{What kind of counterfactual claim is being made?}
CPC begins with the clinical question rather than the model. For trials, CPC registration maps onto ICH E9(R1) estimands \citep{ich2019e9r1}. Before evaluation, the intended population, intervention or comparison, time horizon, outcome and intended patient histories should be stated clearly. This matters because different questions require different evidences. A digital twin that is credible for an average treatment effect is not automatically credible for a patient-specific treatment contrast or a mechanistic explanation.

CPC distinguishes nine formal counterfactual claim classes: outcome (CC-O), trajectory (CC-T), average treatment effect (CC-ATE), conditional average treatment effect (CC-CATE), individual treatment effect (CC-ITE), structural (CC-STR), principal-stratum (CC-PS), attribution (CC-ATTR), and policy-value (CC-V) claims (Supplementary Table S2). These classes capture differences in the target of the claim, its causal structure, temporal form and intervention type. For a clinician-facing description, CPC groups the nine classes into five broader families, reported in Table~\ref{tab:claims_simple}. The class labels are shown here so that the operational taxonomy is visible in the main text; target functionals, notation and identification requirements are retained in Supplementary Notes~1--2.

\begin{table*}[t]
\centering
\caption{ The CPC claim families. The same digital twin may support several claim types, but each must be evaluated separately.}
\label{tab:claims_simple}
\small
\begin{tabularx}{\textwidth}{p{3.5cm}p{4.3cm}X}
\toprule
Claim family & Clinical question & Main evidential boundary \\
\midrule
Intervention response

{\scriptsize ($\mathsf{CC}$-O, $\mathsf{CC}$-T)} & What would happen under a specified treatment or treatment policy? & Requires a well-defined intervention, adequate data support and credible handling of longitudinal observation and treatment history. \\
Population effect

{\scriptsize ($\mathsf{CC}$-ATE, $\mathsf{CC}$-CATE)} & How would two interventions differ on average, or within a defined subgroup? & Population or subgroup effects may be estimable or bounded, but they do not reveal the missing outcome for a particular patient. \\
Individual counterfactual 

{\scriptsize ($\mathsf{CC}$-ITE)} & What is the realized difference for patient $i$ under two interventions? & The expected contrast for patients sharing $i$'s history is CC-CATE; a realized CC-ITE needs cross-world assumptions. \\
Structural or attributional

{\scriptsize ($\mathsf{CC}$-STR, $\mathsf{CC}$-PS, $\mathsf{CC}$-ATTR)} & Through which pathway, mechanism or subgroup structure would an effect arise? & Mechanistic or attributional conclusions require extra structural and pathway assumptions beyond those needed for a population effect. \\
Decision or policy 

{\scriptsize ($\mathsf{CC}$-V)} & Which treatment policy should be preferred? & Requires evidence for the relevant treatment histories, outcomes and clinical utility over the population where the policy will be used. \\
\bottomrule
\end{tabularx}
\end{table*}

Several causal requirements recur across the five claim families. The intervention must be defined precisely enough that the observed and hypothetical treatments refer to the same treatment version, timing and adherence conditions. Patients receiving different treatments must be sufficiently comparable after accounting for relevant history; the required treatment options must be represented in the data over the histories where the claim is made; and evidence transferred across settings must be applicable to the target population \citep{cole2009consistency,robins1986gformula,robins2000msm,petersen2012diagnosing,pearl2022external,dahabreh2019generalizing}. These conditions correspond to consistency, exchangeability, positivity and transportability assumptions. They cannot simply be assumed to hold because a model predicts well.

A key limitation arises when moving from population-level treatment effects to counterfactual outcomes for an individual patient. Under suitable causal assumptions, real-world or trial data can sometimes identify or bound intervention-specific risks and average treatment contrasts. Those quantities do not generally determine the paired outcomes that the same individual would experience under two mutually exclusive treatments \citep{manski1990bounds,tian2000probabilities,fan2010sharp,zhang2025itebounds}. Individual and certain structural or attributional claims therefore require additional cross-world or structural assumptions. If those assumptions are not defensible, the claim should be restricted, subjected to sensitivity or bounding analysis, or reported as Indeterminate rather than established. An individualized conditional average remains CC-CATE; it is not the realized patient-specific contrast CC-ITE.

A concrete example makes the distinction between population-level and patient-specific claims immediately apparent. Consider a digital twin used to study a fixed statin intervention versus a comparator in dementia. Asking for the average five-year difference in dementia risk is a population-effect claim. Asking for that patient's realized difference is CC-ITE; asking for the conditional average given the patient's history is CC-CATE. Asking whether any difference operates through a cardiovascular pathway is a structural claim. Treatment selection generally relies on such conditional averages, not realized ITEs; these claims carry different evidential requirements. The example is illustrative only and does not assert a clinical effect of statin on dementia.

\section{The Three-Reality Concordance Framework}
\begin{figure*}[t]
    \centering
    \includegraphics[width=0.8\linewidth]{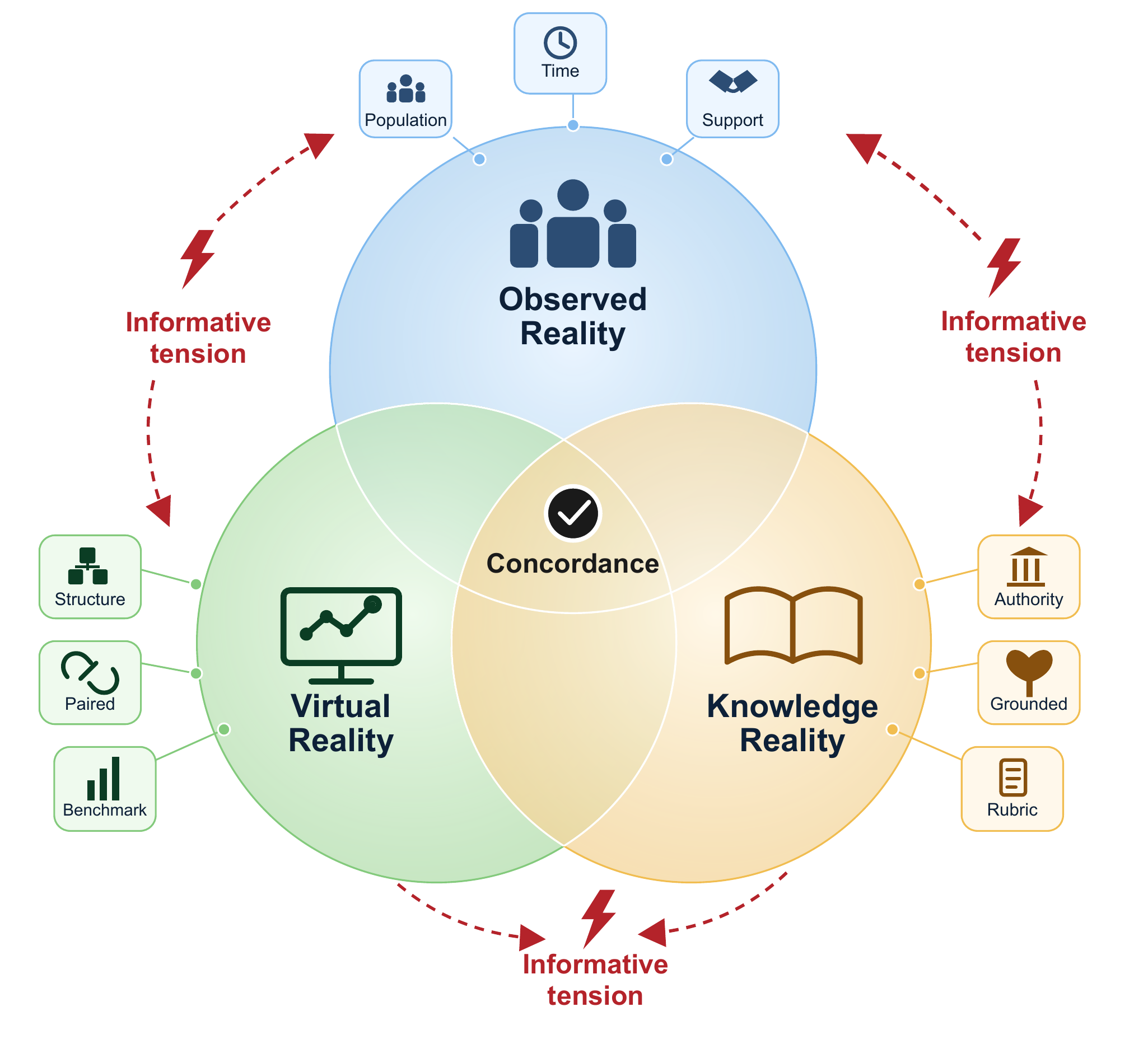}
    \caption{\textbf{The Three-Reality Concordance Framework.} Observed Reality asks whether the claim is compatible with real-world evidence; Virtual Reality (qualified-simulation evidence) tests recovery in declared simulations; Knowledge Reality asks whether the result is clinically and mechanistically admissible. Compatibility strengthens claim-specific credibility only within each reality's stated assumptions and authority. The surrounding labels denote population, time, support, structure, paired testing, benchmarking, authority, grounding and adjudication rubric.}
    \label{fig:realities}
\end{figure*}

CPC organizes evidence into three complementary realities, illustrated in Fig.~\ref{fig:realities}. None is counterfactual ground truth.

\textbf{Observed Reality: what happened in patients?} Real-world longitudinal data anchor factual behavior and can also constrain intervention effects when the causal assumptions are defensible. Relevant tools include target-trial emulation, causal estimates, sensitivity analyses, partial-identification bounds, transport analyses and negative controls \citep{keogh2024interventions,boyer2025counterfactual,hernan2016targettrial,lipsitch2010negative}. Observed Reality remains unable to reveal both alternative outcomes for the same patient.

\textbf{Virtual Reality: what could happen under the declared model?} A qualified simulator can create the same virtual patient under alternative interventions and expose paired outcomes that cannot be observed in real patients. This provides a direct test of whether the IC-HDT recovers counterfactual behavior \emph{within that simulator}. Such recovery is useful but conditional: it does not establish that the simulator's paired outcomes are the true clinical counterfactual.

\textbf{Knowledge Reality: what is clinically admissible?} Clinicians, guidelines, systematic reviews, pharmacological resources, ontologies and biomedical knowledge graphs can identify trajectories that are clinically, temporally, pharmacologically or mechanistically implausible. They cannot provide the missing potential outcome. A clinician-calibrated medical large language model may help organize or apply this evidence, but only as a grounded interface to qualified biomedical knowledge; it is not a fourth reality or an independent source of truth.

\begin{table*}[t]
\centering
\caption{What each CPC reality contributes and where its authority ends.}
\label{tab:qualification_simple}
\small
\begin{tabularx}{\textwidth}{p{2.4cm}p{4.4cm}X}
\toprule
Reality & Useful contribution & Limit that must remain explicit \\
\midrule
Observed & Factual performance, real-world causal estimates or bounds, support and transport checks, and negative controls. & Cannot reveal the missing paired outcome for the same real patient. \\
Virtual & Direct recovery tests for paired outcomes, trajectories or pathways under a declared simulator and intervention structure. & Demonstrates compatibility with that simulator, not clinical counterfactual truth. \\
Knowledge & Clinical, temporal, pharmacological and mechanistic admissibility based on qualified external evidence and expert review. & Can reject implausible behavior but cannot supply the missing counterfactual outcome. \\
\bottomrule
\end{tabularx}
\end{table*}

For CPC use, each Reality must also meet minimum qualification standards. Observed-Reality evidence requires target-population alignment, a well-defined intervention, temporal validity, adequate treatment support and a defensible causal analysis. Virtual-Reality evidence requires transparent and versioned causal structure, modular interventions, an exposed paired counterfactual benchmark where required by the claim, perturbable assumptions and reproducible stochastic evaluation. Knowledge-Reality evidence requires authoritative and versioned sources, a clinician-defined adjudication rubric, robustness testing, explicit abstention and escalation rules, and auditable provenance. 
Concordance denotes structured compatibility across the available evidence, not a single numerical score or observation of the counterfactual. Disagreement between the evidence from the three realities is also informative. Observed--Virtual disagreement may indicate simulator misspecification, population mismatch or unresolved confounding. Observed--Knowledge disagreement may reveal measurement problems, context differences or limits of current knowledge. Virtual--Knowledge disagreement may expose implausible simulator mechanisms. Disagreement warrants investigation; Fail requires a qualified, prespecified mandatory contradiction, while unresolved disputes remain Indeterminate. Clinical knowledge is authoritative but revisable, not ground truth. Cross-reality disagreements are subject to structured investigation and adjudication, with findings informing subsequent model refinement and versioned re-evaluation (Supplementary Note~3).

Apparent agreement can also be misleading when the realities share the same data, assumptions or source material. For example, a simulator calibrated on the evaluation cohort and a knowledge source dominated by studies from the same cohort do not provide three independent confirmations. CPC therefore records evidence provenance and asks whether the conclusion remains stable when important shared sources or assumptions are removed or varied; shared identification assumptions cannot be made independent by this audit. Detailed qualification standards for all three realities are provided in Supplementary Note~4.

An optional clinician-calibrated medical large language model can apply qualified sources, but never supply independent evidence; unsafe or unstable results require human review (Supplementary Note~3) \citep{croxford2025judge,tam2024quest}.

\section{From concordance to a CPC decision}
CPC turns three-Reality Concordance Framework ideas into a six-stage claim-level workflow (Fig.~\ref{fig:cpc}). The essential decision rule is summarized below; the full operationalisation is retained in Supplementary Note~3.

\begin{figure*}[t]
    \centering
    \includegraphics[width=\linewidth]{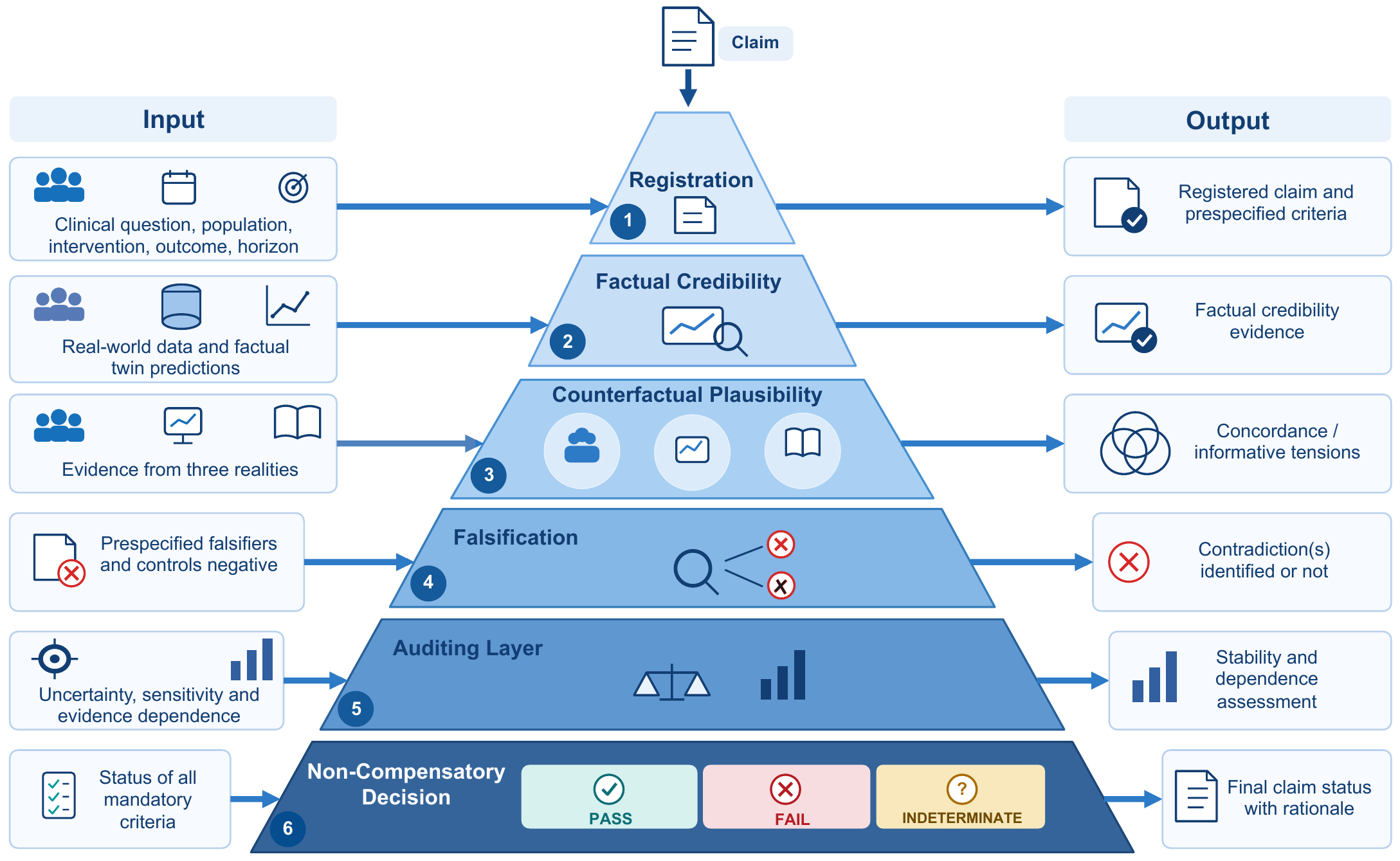}
    \caption{\textbf{CPC claim-level evaluation workflow.} Evaluation begins by defining the clinical question and context of use. The claim is then checked for factual credibility, counterfactual plausibility, contradiction, uncertainty and evidence dependence before a final Pass, Fail or Indeterminate status is reported.}
    \label{fig:cpc}
\end{figure*}

\textbf{1. Register the claim and its context of use.} Before evaluation, the assessor should specify the target population and relevant subgroups, intervention or comparison, outcome, horizon, intended histories, causal assumptions and criteria that could alter the decision. The multidisciplinary evaluation team should prespecify the minimum claim-class checks, thresholds, discrepancy tolerances and sensitivity ranges using clinically meaningful margins and available independent evidence; these must not be retuned after evaluation. Unresolved threshold uncertainty or instability yields Indeterminate for the affected mandatory check. Registration and stipulated verdicts are illustrated in Supplementary Note~5.

\textbf{2. Establish factual credibility.} The twin should first reproduce clinically relevant aspects of the observed course under the treatment history that actually occurred. This includes prediction, calibration, distributional and temporal behavior, and uncertainty. The analysis should account for selective observation caused by censoring, irregular visits, dropout, switching, competing events or missing measurements. A factual pass is necessary but never sufficient for a counterfactual pass: reproducing what happened does not establish what would have happened under another treatment. Failure of a registered mandatory factual criterion yields Fail; factual inadequacy outside that criterion yields Indeterminate when it undermines claim support (Supplementary Note~3).

\textbf{3. Assess counterfactual plausibility with claim-matched evidence.} Observed Reality should be used to constrain the claim as far as the available causal evidence allows---for example through a trial estimate, target-trial analysis, sensitivity region or partial-identification bound. Compatibility alone is insufficient for Pass when the evidence gives an uninformative range; CPC preserves that uncertainty \citep{imbens2004partial,stoye2009partial}. Virtual Reality is most useful for paired or pathway information unavailable in patients; recovery remains conditional on simulator structure and cross-world coupling, which must be varied where relevant. Knowledge Reality tests whether the generated behavior is clinically admissible. These sources are complementary, not interchangeable: simulation cannot repair missing clinical identification, and clinical plausibility cannot replace causal support.

\textbf{4. Search for contradictions through prespecified falsification.} CPC deliberately looks for ways the claim could fail. Negative controls, violated observable implications, null-effect simulation scenarios, perturbations of simulator assumptions, or impossible clinical trajectories can all act as falsification tests~\cite{cornish2026causal,lipsitch2010negative}. A falsifier earns P only if a suitably powered registered test finds no contradiction; P denotes test completion, not positive causal support. Otherwise the check is Indeterminate or Fail (Supplementary Note~3).

\textbf{5. Audit uncertainty and evidence dependence.} The evaluator should examine whether positive checks share data, calibration targets, assumptions, literature or experts; whether the judgment is stable across prespecified uncertainty ranges; and whether support differs across relevant subgroups~\cite{lawlor2016triangulation,sel2025vvuq,paulus2020predictably}. If removing or perturbing a shared lineage changes the judgment, that dependence must be reported; if no usable evidence remains, the affected check is Indeterminate rather than independently confirmed. Material unresolved dependence or instability blocks Pass and yields Indeterminate unless a registered mandatory contradiction already establishes Fail.

\textbf{6. Report Pass, Fail or Indeterminate.} CPC is deliberately non-compensatory. For mandatory checks $\ell\in\mathcal L_c$, let $G_\ell^c\in\{\mathrm P,\mathrm F,\mathrm I\}$ denote Pass, Fail or Indeterminate. The claim-level decision is
\begin{equation}
G_{\mathrm{CPC}}(\mathsf{CC}_c)=
\begin{cases}
\mathrm F, & \text{some mandatory check is F},\\
\mathrm I, & \text{none is F and some mandatory check is I},\\
\mathrm P, & \text{all mandatory checks are P}.
\end{cases}
\label{eq:cpc_gate_main}
\end{equation}
where $\mathsf{CC}_c$ is the registered counterfactual claim. Thus, a valid contradiction to any mandatory criterion produces \textbf{Fail}; strong performance elsewhere cannot cancel it. If a required assumption, evidence source, support condition or audit remains unresolved, the result is \textbf{Indeterminate}. \textbf{Pass} is reserved for claims for which all mandatory evidence is compatible with the prespecified requirements, support and context of use. Pass therefore means ``supported by the available evidence under stated assumptions at the time of evaluation,'' not ``the counterfactual is true.'' Indeterminate reasons should be recorded separately, distinguishing missing evidence, uncertain assumptions, unresolved disagreement, inadequate subgroup support and instability. Indeterminate reasons are non-exclusive: I-E, insufficient evidence; I-A, unresolved assumptions; I-D, evidence disagreement; I-S, inadequate support or transportability; and I-U, sensitivity or provenance instability. An indeterminate result can identify the evidence or adjudication needed for re-evaluation and help guide future trial design (Supplementary Note~3).

The distinction between population-level treatment effects and patient-specific counterfactual claims is clearest for an individual treatment question. Real-world data may constrain what tends to happen under each treatment across comparable patients, and a digital twin may reproduce the correct average treatment effect. Neither fact reveals the two outcomes that the same patient would have experienced under both treatments. Simulated recovery and clinical admissibility cannot replace unsupported cross-world assumptions; the realized individual claim then remains Indeterminate.

\section{Discussion}
CPC is a methodological Perspective, not an empirical validation study. Its purpose is to make the evaluation of counterfactual digital-twin claims more explicit, clinically interpretable and difficult to overstate. Three consequences follow.

First, a digital twin should not receive one global credibility label. Credibility belongs to a specific claim in a specific context of use~\cite{burr2024trustworthy}. The same system may support a population effect while remaining Indeterminate for a patient-specific or inadequately represented subgroup claim. A pooled Pass does not validate an unsupported subgroup or establish fairness~\citep{liu2023translational}. Different stakeholders may legitimately interpret CPC verdicts differently, so verdict communication should specify intended use and user.

Second, different claims require different evidence. Real-world data can assess factual behaviour and constrain causal claims, but cannot reveal both potential outcomes for the same patient. Simulation can expose paired outcomes only within its declared causal structure. Clinical and mechanistic knowledge can reject implausible behaviour without establishing counterfactual truth. These evidence streams are complementary, not interchangeable.

Third, agreement between evaluation checks does not necessarily provide independent evidence. Different checks may rely on the same dataset, assumptions or clinical knowledge, so they may reach similar conclusions for the same underlying reason. Credibility assessment should therefore record where each source of evidence comes from and examine whether the conclusion changes when shared sources are varied or removed. This is particularly important when the same information is used to calibrate, evaluate and clinically interpret a digital twin.

The next step for CPC is empirical testing rather than adding further conceptual complexity. Controlled simulation studies should measure false-pass and false-fail behavior as confounding, treatment support, heterogeneity, longitudinal feedback, transport shift and simulator structure are varied. Semi-real or real studies should compare population-level claims against randomized evidence, credible target-trial analyses or policy changes where such comparisons are possible. Clinical adjudication studies should test Knowledge-Reality judgments under deliberately seeded clinical and mechanistic violations.

Several limitations cannot be eliminated by the proposed framework. Individual counterfactual effects may not be fully identifiable from the available data, and simulation-based evidence remains dependent on the assumptions and structure of the simulator. Conclusions may also fail to generalize to populations, sites or subgroups that are poorly represented in the available evidence, while external biomedical knowledge may change over time. Indeterminate must prompt evidence collection, narrower registration or withholding unsupported clinical use; it may be frequent under non-compensatory aggregation.

CPC judgments are conditional and revisable, not statements of counterfactual truth. A Pass or Fail reflects the evidence, simulator, assumptions and clinical knowledge available at evaluation; none can exclude unknown mechanisms or future corrections to biomedical knowledge. Material new evidence, revised clinical constraints, newly identified mechanisms or simulator changes should therefore trigger re-evaluation under the same decision rule (Supplementary Note~3).

The framework does not solve the missing-counterfactual problem. Its purpose is narrower and more practical: to prevent evidence for one kind of claim from being used to justify a stronger one. The central discipline is unchanged: \textbf{evidence strength must match claim strength, and concordance must never be confused with counterfactual truth}.

\section*{Data availability}
No new patient-level data were generated or analyzed for this methodological Perspective.

\section*{Code availability}
No new code accompanies this Perspective. Future CPC implementations should release claim-registration schemas, evaluation code, simulator configurations, falsification queries, adjudication rubrics, retrieval-corpus versions and evidence-provenance records sufficient for reproduction.

\section*{Acknowledgements}
P.E. is supported by the Medical Research Council (MRC; MR/W025728/1) and the National Institute for Health and Care Research (NIHR) Southampton Biomedical Research Centre (BRC; NIHR203319). N.C.H. is supported by the MRC (MC\_PC\_21003 and MC\_PC\_21001), the NIHR Southampton BRC (NIHR203319), and an NIHR Senior Investigator Award (NIHR305844). S.M. is supported by the NIHR (NIHR306220). R.A. is supported through the NIHR Southampton BRC (NIHR203319).


\section*{Author contributions}
S.F. and R.A. contributed to conceptualization, methodology, investigation, visualization, writing of the original draft, and review and editing of the manuscript. R.A. additionally contributed funding acquisition, project administration, and supervision. C.K., J.M., G.C., S.M., B.M., S.E., P.E. and N.C.H. contributed to the critical evaluation and refinement of the methodology, clinical and scientific interpretation, and substantive review and editing of the manuscript. All authors reviewed and approved the final manuscript.

\section*{Competing interests}
The authors declare no competing financial or non-financial interests.

\section*{Use of generative AI}
During the preparation of this manuscript, the authors
used generative AI tools  to assist with language
editing and improving the clarity and readability of
the text. All AI-assisted modifications were reviewed
and verified by the authors, who take full responsibility
for the final content of the manuscript.

\FloatBarrier
\printbibliography

\end{document}


\maketitle

\section*{Supplementary Note 1: Technical notation and identification requirements}
For patient $i$, $X_{ik}=\langle M_{ik},A_{ik},Y_{ik}\rangle$ denotes the recorded state at visit $k$ and $H_{ik}=(F_{ik},s_i,\Delta t_{i,1:k})$ the recorded history, with $F_{ik}=\{X_{i1},\ldots,X_{ik}\}$. A dynamic policy $\pi_k(a\mid H_{k-1})$ may depend on evolving history \citep{murphy2003dtr,robins2000msm}. We use $g$ for a generic fixed or policy-based intervention and $X^g$ for its potential trajectory.

For a CPC claim $c$, the formal claim specification is
\begin{equation}
\CC_c=(\mathcal P_c,\mathcal G_c,\tau_c,\psi_c;\Omega_c),
\label{eq:s_claim}
\end{equation}
where $\mathcal P_c$ is the target population, $\mathcal G_c$ the intervention or intervention contrast, $\tau_c$ the horizon, $\psi_c$ the causal quantity being claimed, and $\Omega_c$ the registered intended region of patient histories. Empirical support is assessed against this region; restricting to supported histories narrows and re-registers the claim rather than retroactively passing the original claim. This parallels estimand specification for clinical trials, including treatment handling and intercurrent events \citep{ich2019e9r1}, while covering broader counterfactual claims. It records who, which intervention, when, which causal quantity and over which intended histories.

For a history-dependent policy $\pi$, $A_k^{\pi}(H_{k-1})$ denotes the action prescribed at decision time $k$. Here, $b$ denotes the intervention branching time. Consistency requires the observed action to match that prescription at every relevant decision time:
\begin{equation}
\{A_k=A_k^{\pi}(H_{k-1})\text{ for all }k=b+1,\ldots,t\}
\Longrightarrow X_{b+1:t}=X_{b+1:t}^{\pi}.
\end{equation}
Sequential exchangeability may be represented as
\begin{equation}
A_k\perp\!\!\!\perp (M_{k:t}^{\pi},Y_{k:t}^{\pi})\mid H_{k-1},
\end{equation}
and sequential positivity is assessed for every registered evolving history $H_{k-1}\in\Omega_{c,k-1}$,
\begin{equation}
\pi_k(a\mid H_{k-1})>0\Longrightarrow
\Pp(A_k=a\mid H_{k-1})>0.
\end{equation}

The plain-language causal requirements in the main text map directly to the following formal identification requirements. Consistency, exchangeability, positivity and transportability correspond to A1--A4; the additional assumptions required for individual and structural/pathway claims correspond to A5--A6. 

The CPC uses the following identification requirements. \textbf{A1, consistency} \citep{cole2009consistency}: the observed outcome equals the relevant potential outcome when the observed action matches the registered intervention, including treatment version, timing, dose and adherence. \textbf{A2, sequential exchangeability} \citep{robins1986gformula,robins2000msm,murphy2003dtr}: conditional on recorded history, treatment is independent of future potential outcomes; this is not testable from observed data alone. \textbf{A3, sequential positivity} \citep{robins1986gformula,petersen2012diagnosing}: every action required by the registered intervention or policy has non-zero probability over the intended histories; violations require a narrowed claim or Indeterminate. \textbf{A4, transportability} \citep{pearl2022external,dahabreh2019generalizing}: when a claim is transferred across populations, relevant causal relations must be transportable after conditioning on effect modifiers and with adequate covariate overlap. \textbf{A5, cross-world dependence}: joint, individual, principal-stratum and attribution quantities require an explicit cross-world dependence model or structural restriction; without it they are generally only partially identified \citep{manski1990bounds,tian2000probabilities,fan2010sharp,zhang2025itebounds}. \textbf{A6, pathway identification}: structural claims require mediator-specific temporal, confounding, intervention and support assumptions; natural effects can impose additional cross-world restrictions.

Three levels should be distinguished. The marginal laws $P(Y^{g_1})$ and $P(Y^{g_0})$ may sometimes be identified under A1--A4; those marginals do not generally identify the joint law $P(Y^{g_1},Y^{g_0})$; and the joint law does not make the realized individual contrast $Y_i^{g_1}-Y_i^{g_0}$ observable. For claims requiring A5/A6, default to I-A (unresolved assumptions) unless a registered clinical-methodological panel (including clinical and causal-inference expertise independent of model development where feasible) documents defensible restrictions and sharp bounds where available; any P is explicitly conditional on those restrictions. Unsupported assumptions require narrower claims, bounds or Indeterminate.

For a generic paired causal simulator, $g_M$, $g_Y$ and $g_A$ denote mediator/state, outcome and treatment mechanisms; $\delta t_k$ is elapsed time; and $U_{Mk}$, $U_{Yk}$ and $U_{Ak}$ are declared exogenous stochastic terms:
\begin{equation}
M_k=g_M(H_{k-1},A_k,\delta t_k,U_{Mk}),\quad
Y_k=g_Y(H_{k-1},M_k,A_k,\delta t_k,U_{Yk}),\quad
A_k=g_A(H_{k-1},U_{Ak}).
\end{equation}
Branching the same virtual patient under interventions $g_0,g_1$ while preserving the declared cross-world coupling exposes simulator-defined contrasts such as
\begin{equation}
\Delta_{i,q}^{(s)}=Y_{i,b+q}^{g_1,s}-Y_{i,b+q}^{g_0,s}.
\end{equation}
Here $i$ indexes the virtual patient, $q$ the post-branching horizon and $s$ the simulator. These quantities are benchmarks conditional on simulator structure and noise coupling, not clinical counterfactual ground truth.

\paragraph{Methodological foundations and their integration into CPC.}

CPC builds on established contributions to model credibility,
clinical-claim-based validation, causal falsification and
counterfactual identification. These complementary foundations
inform CPC's integrated, operationally specified framework,
in which claim-specific causal assumptions, qualified evidence
streams and falsification checks are linked to auditable,
non-compensatory decisions. Table~\ref{tab:s_prior_approaches}
illustrates how these contributions are incorporated into
CPC's evaluation procedure.

\begin{table}[H]
\centering
\caption{Established methodological foundations and their
distinct roles within the proposed CPC evaluation framework.
CPC builds on these approaches by linking claim-specific
causal requirements, qualified evidence and falsification
to auditable, non-compensatory decisions.}
\label{tab:s_prior_approaches}

\scriptsize
\setlength{\tabcolsep}{3pt}
\renewcommand{\arraystretch}{1.15}

\begin{tabularx}{\textwidth}{
@{}p{2.55cm}p{3.2cm}p{3.45cm}X@{}
}
\toprule

Prior approach &
Established contribution &
Elements adopted in CPC &
Additional operational role within CPC \\

\midrule

ASME V\&V40 \cite{asme2018vv40};
ICH E9(R1) \cite{ich2019e9r1}
&
Risk-informed, context-of-use model credibility;
prespecification of clinical estimands.
&
Registration of the intended population,
intervention, outcome, horizon and context
of use.
&
Links each registered counterfactual claim
to its applicable causal identification
assumptions and mandatory evidence checks.
An unmet mandatory requirement prevents
an unqualified Pass.
\\

\addlinespace[3pt]

Vall\'ee
\cite{vallee2026digital,vallee2026jamia,
vallee2026causaltwins}
&
Clinical-claim-based validation,
decision relevance and guidance
for causal digital twins.
&
Alignment of validation evidence
with the intended clinical claim
and its decision context.
&
Connects nine formal counterfactual
claim classes to claim-specific
identification requirements, qualified
evidence streams and explicit
Pass, Fail or reason-coded
Indeterminate decisions.
\\

\addlinespace[3pt]

Cornish et al.
\cite{cornish2026causal}
&
Causal falsification using
testable implications and
potential contradictions.
&
Prespecified falsification checks
for assessing compatibility
with causal claims.
&
Incorporates qualified contradictions
as non-compensatory Fail triggers
for mandatory criteria.
Inconclusive or insufficiently
informative tests cannot
establish Pass.
\\

\addlinespace[3pt]

Laudy
\cite{laudy2026dtcf}
&
Distinction between identifiable
marginal quantities and
cross-world dependence;
partial identification and bounds.
&
Explicit separation of marginal,
joint and individual
counterfactual quantities.
&
Links identification limitations
to claim-specific decision rules.
Unresolved cross-world assumptions
produce reason-coded Indeterminate
outcomes; simulator-based paired
contrasts do not establish
clinical counterfactual truth.
\\

\bottomrule
\end{tabularx}

\vspace{3pt}
\parbox{\textwidth}{
\footnotesize
\emph{Interpretation.}
The individual methodological principles listed above
are established contributions.
CPC's proposed additional contribution lies in their
operational integration: matching counterfactual claims
to qualified evidence, auditing causal assumptions
and evidence dependencies, and producing explicit,
non-compensatory decisions.
These proposed operational roles do not constitute
independent empirical validation of CPC.
}

\end{table}
\FloatBarrier

\FloatBarrier

\section*{Supplementary Note 2: Full counterfactual claim taxonomy and dementia--statin examples}

The five clinician-facing claim families in the main text retain nine formal counterfactual claim classes: Intervention Response maps to $\CC$-O and $\CC$-T; Population Effect to $\CC$-ATE and $\CC$-CATE; Individual Counterfactual to $\CC$-ITE; Structural or Attributional to $\CC$-STR, $\CC$-PS and $\CC$-ATTR; and Decision or Policy to $\CC$-V. The target functionals for structural, principal-stratum and attribution claims are illustrative rather than exhaustive. $b$ denotes the branching point, $M$ a mediator and $R$ a post-treatment variable.

\begin{table}[h]
\centering
\caption{Full CPC counterfactual claim taxonomy, representative target functionals, principal identification requirements and real-world constraints.}
\label{tab:s1}
\scriptsize
\begin{tabularx}{\textwidth}{p{1.25cm}p{2.9cm}p{3.0cm}X}
\toprule
Class & Representative target functional & Principal requirements & Main real-world constraint / residual limitation \\
\midrule
$\CC$-O & $\E(Y_q^{g})$ or $F_{Y_q^g}$ & A1--A3; A4 when transported & Interventional mean, distribution or bounds; support and identification uncertainty may remain. \\
$\CC$-T & $P(X_{b+1:t}^{g}\mid h)$ & Sequential A1--A3; A4 when transported & Longitudinal interventional distribution; sparse support and long-horizon uncertainty may remain. \\
$\CC$-ATE & $\E(Y^{g_1}-Y^{g_0})$ & A1--A3; A4 when transported & Population causal contrast, sensitivity region or bounds; no individual cross-world effect is implied. \\
$\CC$-CATE & $\E(Y^{g_1}-Y^{g_0}\mid H_b=h)$ & A1--A3; A4 when transported & Conditional causal contrast under local support; sparse histories may remain unsupported. \\
$\CC$-ITE & Realized difference for patient $i$: $Y_i^{g_1}-Y_i^{g_0}$ & A1--A3 for marginals plus A5 & An individualized conditional mean is CC-CATE, not CC-ITE; joint dependence requires A5 and a realized individual contrast remains unobserved. \\
$\CC$-STR & $\E[Y^{a_1,M^{a_1}}-Y^{a_1,M^{a_0}}]$ & A1--A3 plus A6; A5 when cross-world contrasts are used & Pathway-specific constraints; the asserted pathway remains assumption-dependent. \\
$\CC$-PS & $\E[Y^{a_1}-Y^{a_0}\mid(R^{a_1},R^{a_0})=r]$ & A1--A3 plus A5; additional restrictions when required & Bounds or partially identified principal-stratum effects; stratum membership and cross-world dependence remain unobserved. \\
$\CC$-ATTR & $\Pr(Y^{a_0}=0\mid Y^{a_1}=1)$ & A1--A3 plus A5; attribution assumptions as required & Attribution bounds, sensitivity regions or identified quantities under additional assumptions; cross-world dependence generally remains unidentified. \\
$\CC$-V & $V(\pi)=\E\{u(X^\pi)\}$ & Sequential A1--A3, policy support and utility; A4 when transported & Policy value, regret or bounds over supported histories; unsupported policies or histories remain Indeterminate. \\
\bottomrule
\end{tabularx}
\end{table}

For illustration only, let fixed statin and comparator interventions be $a_S$ and $a_0$, and history-dependent policies be $\pi_S$ and $\pi_0$. These questions do not assert a statin effect on dementia or cardiovascular mediation.

\begin{table}[h]
\centering
\caption{Toy dementia--statin questions illustrating the nine formal claim classes.}
\label{tab:s2}
\scriptsize
\begin{tabularx}{\textwidth}{p{1.3cm}X p{3.8cm}}
\toprule
Claim & Counterfactual toy question & Domain / classification \\
\midrule
$\CC$-O & What would five-year dementia risk be under fixed statin intervention $a_S$? & Intervention Response; population; single-point. \\
$\CC$-T & How would cognition, cardiovascular state and dementia status evolve under $a_S$? & Intervention Response; personalized; longitudinal. \\
$\CC$-ATE & What is the average five-year risk difference under $a_S$ versus $a_0$? & Population Effect; population; single-point. \\
$\CC$-CATE & What is the average contrast under $a_S$ versus $a_0$ among histories $H_b=h$? & Population Effect; conditional population/subgroup. \\
$\CC$-ITE & What is the realized five-year difference for patient $i$ under $a_S$ versus $a_0$ (not the CC-CATE expected difference for histories like $i$'s)? & Individual Counterfactual; personalized. \\
$\CC$-STR & How much of the fixed-intervention contrast is attributed to $A\rightarrow M^{\mathrm{CV}}\rightarrow Y$? & Structural and Attributional; structural. \\
$\CC$-PS & Among patients with good cardiovascular status at two years under both $a_S$ and $a_0$, what is the average five-year dementia-risk difference? & Structural and Attributional; principal stratum. \\
$\CC$-ATTR & Among patients who would develop dementia within five years under $a_S$, what proportion would not under $a_0$? & Structural and Attributional; attribution. \\
$\CC$-V & Which statin-initiation policy, $\pi_S$ or $\pi_0$, yields greater expected utility? & Decision and Policy; longitudinal policy. \\
\bottomrule
\end{tabularx}
\end{table}

\FloatBarrier
\section*{Supplementary Note 3: Quantitative CPC operationalisation}
The reader-facing workflow separates evidence about the observed course from evidence constraining the unobserved alternative. Formally,
\begin{equation}
\CC_c\rightsquigarrow(\CC_c^{F},\CC_c^{CF}),\qquad
\CPC(\CC_c)=\bigl(\FC(\CC_c^{F}),\CFC(\CC_c^{CF})\bigr),
\label{eq:s_cpc}
\end{equation}
where $\CC_c^{F}$ is the factual projection, $\CC_c^{CF}$ the counterfactual component, and $\FC$ and $\CFC$ denote the corresponding evidential evaluations. This is an evidential decomposition, not an algebraic decomposition of the target causal quantity. These evidential evaluations inform the mandatory checks whose grades are combined into one CPC verdict in Stage 6.

\subsection*{Stage 1: Claim registration}
Stage 1 specifies three objects. The \emph{claim specification} records $(\mathcal P_c,\mathcal G_c,\tau_c,\psi_c,\Omega_c)$, the claim class/domain, identification-assumption set $\mathcal A_c$, context of use $\chi_c$ and prespecified relevant subgroups.

The \emph{evidence specification} records mandatory checks $\mathcal L_c$, safety-critical checks and falsifiers, factual-statistic family $\Phi_F^c$, real-world causal constraint and sensitivity specification $\Gamma$, paired-simulator family and cross-world coupling, Knowledge-Reality sources, and provenance/dependence plan.

The \emph{decision specification} records discrepancy measures, scaling rules, thresholds, any clinical-indifference threshold, calibration protocol, multiplicity control and optional adjudication rules.

Every claim requires factual, Observed-Reality, Knowledge-Reality, falsification and provenance checks; unavailable evidence receives I, not an omission. Virtual recovery is mandatory for CC-ITE, CC-PS, CC-ATTR and cross-world CC-STR, and otherwise if paired/pathway evidence is invoked. The intended region is fixed at registration. Requirements and thresholds are proportional to clinical model risk $\chi_c$; exclusions or absent checks are documented and cannot silently produce Pass. Prespecified falsification thresholds, discrepancy tolerances and sensitivity ranges are set before final evaluation by the registered multidisciplinary evaluation team, including relevant clinical or domain expertise and, where feasible, evaluators independent of model development. Their values and plausible ranges are justified using clinically meaningful margins or safety requirements, independent calibration/reference data, reference distributions or benchmark cases, and external evidence for non-identifiable sensitivity parameters. The responsible authority, evidence source, calibration protocol and plausible range are recorded. Thresholds are not retuned after inspecting final evaluation results; status instability across the registered plausible range, or absence of a defensible range, yields Indeterminate for the affected mandatory criterion.

\subsection*{Stage 2: Factual credibility}
Let $\Theta$ denote the twin model, $\bar A$ the observed
treatment history, and $\phi$ a scalar-valued statistic. The operators $\widehat{\E}_{\Theta}$ and $\widehat{\E}_{\mathrm{RW}}$ denote empirical averages over twin-generated and real-world observations, respectively. For a prespecified family $\Phi_F^c$ of factual statistics, a possible maximum standardized discrepancy is
\begin{equation}
D_F(\CC_c)=\sup_{\phi\in\Phi_F^c}
\frac{\left|\widehat{\E}_{\Theta}\{\phi(\widehat X^{\bar A})\}-\widehat{\E}_{\mathrm{RW}}\{\phi(X)\}\right|}
{\widehat{\sigma}_\phi+\epsilon_0},
\label{eq:s_factual}
\end{equation}
where $\widehat X^{\bar A}$ is generated under the observed action history $\bar A$. Here $\widehat\sigma_\phi$ scales the discrepancy; it is not the standard error of a difference of means. Let $q_{F,c}$ be the registered multiplicity-aware quantile of the reference distribution of $D_F$ under factual compatibility and $U_{F,c}$ a registered upper confidence bound on $D_F$. Under marked skewness or heavy tails, $\widehat\sigma_\phi$ may use a robust scale such as $1.4826\,\mathrm{MAD}\{\phi(X)\}$; rare-event or discrete statistics should use prespecified exact or variance-stabilized alternatives. The numerical floor $\epsilon_0>0$ is not a plausibility tolerance.

For registered mandatory factual targets: $D_F>q_{F,c}$ yields F; $D_F\le q_{F,c}$ and $U_{F,c}\le\varepsilon_{F,c}$ (registered standardized margin corresponding to a clinically meaningful tolerance) yield P; otherwise I-E. Inadequate factual performance outside the registered target is I-E when it undermines the claim, not F. Prespecify observation, censoring, switching, competing-event and missingness handling. Factual P is necessary but insufficient for counterfactual P.

\subsection*{Stage 3: Counterfactual plausibility}
\paragraph{Observed-Reality causal compatibility.}
Let $\widehat\psi_{\Theta,c}$ denote the twin's estimate of
the registered causal quantity $\psi_c$, and
$\mathcal I_{\mathrm{RW}}(\psi_c;\Gamma)$ the real-world
identified or sensitivity set under specification $\Gamma$.
The operator $\mathcal C_{1-\alpha_{\mathrm{RW}}}$
constructs a confidence region at level
$1-\alpha_{\mathrm{RW}}$, while $\operatorname{dist}$
and $\varepsilon_{\mathrm{RW},c}$ denote the prespecified
distance metric and compatibility tolerance.
Observed Reality defines an assumption-indexed compatibility region rather than revealing the missing alternative. When point identification is not justified, compatibility should account for uncertainty in the estimated constraint set:
\begin{equation}
\dist\!\left(\widehat\psi_{\Theta,c},
\mathcal C_{1-\alpha_{\mathrm{RW}}}\!\left[\Iset_{\mathrm{RW}}(\psi_c;\Gamma)\right]\right)
\le \varepsilon_{\mathrm{RW},c},
\label{eq:s_rwcompat}
\end{equation}
where $\mathcal C_{1-\alpha_{\mathrm{RW}}}[\Iset_{\mathrm{RW}}]$ is a confidence region for the identified or sensitivity set rather than a plug-in set treated as known \citep{imbens2004partial,stoye2009partial}. The region may be based on a point estimate with uncertainty, sensitivity region, partial-identification interval, target-trial estimate or transport-adjusted set. Uncertainty in the twin estimate
$\widehat\psi_{\Theta,c}$ must also be accounted for
in the registered compatibility assessment; the displayed point-estimate inequality alone does not determine the uncertainty-aware verdict. For cross-world quantities, sharp or assumption-indexed bounds on treatment-effect distributions and probabilities of causation can be used where available \citep{tian2000probabilities,fan2010sharp,zhang2025itebounds}. Let $C_c=\mathcal C_{1-\alpha_{\mathrm{RW}}}[\Iset_{\mathrm{RW}}]$ and $C_c^{+\varepsilon}=\{u:\dist(u,C_c)\le\varepsilon_{\mathrm{RW},c}\}$. The registered informativeness criterion is $C_c^{+\varepsilon}\cap\mathcal B_c=\varnothing$, where $\mathcal B_c$ is the prespecified competing, decision-relevant conclusion; an optional width limit is evaluated on $C_c^{+\varepsilon}$, not $C_c$. Qualified incompatibility is F; compatibility plus informativeness and defensible identification permits P; uninformative evidence is I-E, doubtful A2 is I-A, and unresolved conflicting analyses are I-D.

\paragraph{Virtual-Reality recovery.}
When a claim requires paired, joint, individual or pathwise information unavailable in observed patients, a qualified causal simulator branches the same virtual patient under alternative interventions while preserving the declared cross-world coupling. CPC compares the IC-HDT-implied joint distribution, individual contrast, trajectory or pathway with the simulator-defined benchmark using the preregistered discrepancy measure and tolerance. A P check verifies recovery only within that simulator family and coupling assumptions; it is a necessary condition, not positive clinical causal support. Alternative coupling assumptions must be varied when relevant; shared coupling between simulator and twin is recorded as a dependence, not independent confirmation.

\paragraph{Knowledge-Reality admissibility and optional clinician-calibrated medical LLM adjudication.}
External biomedical knowledge provides a rejection-only admissibility screen, not positive causal support. A clinician-calibrated medical large language model, when used, is an interface for applying this knowledge rather than an evidence source. It must operate through retrieval-augmented generation (RAG) over a frozen, versioned and cited medical corpus with a prespecified retrieval configuration; unsupported model-memory assertions are inadmissible. Calibration is performed on held-out clinician-labeled cases using sensitivity, specificity, abstention, clinically weighted loss and a prespecified clinically relevant false-concordance target, where false concordance is a clinician-defined contradiction incorrectly adjudicated as admissible. Robustness is re-evaluated under prespecified prompt wording, presentation-order and retrieval perturbations; materially unstable judgments cannot support Pass. Insufficient or conflicting retrieval, safety-critical disagreement, abstention, or failure to meet the calibrated operating target triggers human adjudication \citep{croxford2025judge,williams2026llmjudge,tam2024quest,gu2026llmjudge,shi2025position,ye2025judgebias}. A medical LLM cannot override a valid mandatory contradiction or contribute an independent Reality.

\paragraph{Cross-reality disagreement and adjudication.}
Disagreement between Observed, Virtual and Knowledge Realities warrants investigation rather than automatic rejection. The registered evaluation team should examine the quality, provenance, assumptions and context-specific applicability of the conflicting evidence, including its clinical consequences. A clinician objection or guideline disagreement alone does not establish Fail: the applicability, evidential strength, temporal relevance and clinical consequence of the constraint must first be reviewed. A confirmed contradiction to a qualified, prespecified mandatory criterion yields Fail; unresolved disputes, including possible errors in evolving guidelines or expert consensus, yield Indeterminate. Knowledge sources and adjudications must remain versioned and reviewable. Where adjudication identifies a model or knowledge-base deficiency, the finding should inform subsequent revision and versioned re-evaluation.

\subsection*{Stage 4: Falsification}
Real-world falsifiers may include negative controls or violated observable implications; simulator falsifiers may include null-effect, coupling-perturbation or endpoint-matched/pathway-mismatched scenarios; clinical falsifiers may include impossible temporal, pharmacological or mechanistic behavior. A prespecified falsifier is informative when it has power at least $1-\beta_c$ against the registered minimum material violation at its multiplicity-adjusted threshold. Such a test is P without a qualifying contradiction, F with a confirmed contradiction and I-E if unavailable or insufficiently powered. P records stress-test coverage, not positive causal evidence. A confirmed unexpected safety-critical contradiction suspends any Pass as I-D; after registration and adjudication in the next evidence version, a qualified mandatory contradiction yields F.

\subsection*{Stage 5: Auditing uncertainty and evidence dependence}
Dependence can be represented by a shared-provenance graph whose evidence nodes are linked when they share data, calibration targets, model assumptions, literature or expert inputs. Dependence is material when removing or perturbing a shared lineage changes a mandatory check or the overall CPC status; this should trigger leave-one-lineage-out or alternative-lineage analysis. If this removes all usable evidence, the affected check is I-E, not independent support. Shared causal-identification assumptions remain shared even across different estimators. The audit also evaluates stability over the prespecified sensitivity and threshold ranges. For relevant demographic, clinical or social subgroups, it records treatment support, representativeness, causal-assumption credibility and material differences in claim status~\citep{paulus2020predictably,liu2023translational}. A pooled Pass must not be generalized to a subgroup lacking adequate evidence; such claims should be restricted or reported Indeterminate. This is an audit of subgroup credibility, not a full algorithmic fairness assessment. Unstable status or dependence that cannot be characterized for the context of use yields Indeterminate.

\subsection*{Stage 6: Non-compensatory decision}
For mandatory checks $\ell\in\mathcal L_c$, the non-compensatory decision is
\begin{equation}
G_{\CPC}(\CC_c)=
\begin{cases}
\mathrm F, & \text{some mandatory check is F},\\
\mathrm I, & \text{none is F and some mandatory check is I},\\
\mathrm P, & \text{all mandatory checks are P}.
\end{cases}
\label{eq:s_gate}
\end{equation}
A confirmed F applies only to a registered mandatory target; other factual inadequacy blocks Pass as I-E when material. Statistical failures use registered family-wise error control for confirmatory checks. The final report should state the claim, context of use, assumptions, support, multiplicity control, evidence provenance, residual uncertainty and reasons for any Indeterminate outcome.

\begin{table}[ht]
\centering
\caption{Claim-specific grading of mandatory checks. P means a check is completed at the registered evidential standard, not clinical counterfactual truth; I codes are non-exclusive. Virtual evidence is a verification-type necessary condition; Knowledge is rejection-only. Only qualified Observed-Reality evidence directly supports a clinical causal contrast.}
\label{tab:s_grading_checks}
\scriptsize
\begin{tabularx}{\textwidth}{p{2.05cm}X X X}
\toprule
Mandatory check & P & F & I (primary code) \\
\midrule
Factual & $D_F\le q_{F,c}$ and $U_{F,c}\le\varepsilon_{F,c}$ & $D_F>q_{F,c}$ for registered target & Weak precision or material unregistered mismatch (I-E) \\
Observed Reality & Compatible with informative, qualified causal constraint and defensible assumptions & Qualified, registered causal contradiction & Uninformative (I-E), A2 uncertain (I-A), disagreement (I-D), no support (I-S) \\
Knowledge & Qualified adjudication finds no registered clinical violation & Confirmed registered admissibility violation & Source gap (I-E), unresolved clinical dispute (I-D) \\
Virtual (when required) & Qualified simulator recovery under registered scenarios/couplings & Confirmed registered failure to recover simulator benchmark & Poor qualification/coverage (I-E), coupling instability (I-U) \\
Falsification & Adequately powered registered tests find no contradiction & Confirmed registered falsifier detects material contradiction & Power below $1-\beta_c$ or missing test (I-E) \\
Provenance audit & Dependence and sensitivity audited; required status is stable & Confirmed registered mandatory violation & Empty independent lineage (I-E), unresolved source conflict (I-D), instability (I-U) \\
\bottomrule
\end{tabularx}
\end{table}

\paragraph{Indeterminate reason codes.}
Indeterminate remains a single CPC status with non-exclusive reason codes identifying the unresolved criteria. Absence of a detected contradiction is not evidence of counterfactual truth: P for a qualified falsification check records only adequate stress-test coverage. Indeterminate may be frequent under conjunction; report the unresolved checks and require evidence collection, re-registration to a narrower claim, sensitivity analysis or withholding unsupported clinical use. Unknown unknowns are not a separately detectable category: unrecognized biological mechanisms or other missing knowledge may remain even after Pass.

\paragraph{Versioned re-evaluation.}
The decision in Eq.~\eqref{eq:s_gate} is conditional on the registered evidence state at time $t$ (observational evidence, simulator version, clinical knowledge, assumptions and context of use). Pass means evidential sufficiency under that state, while Fail indicates a qualified registered mandatory contradiction under that state; neither establishes timeless truth. The report should retain source and model versions, review date, responsible adjudicators and re-evaluation triggers. Material new data, a revised guideline or adjudication, an identified mechanism, altered population support or a simulator update require a new assessment using the unchanged non-compensatory rule. A previous Pass or Fail may accordingly change without retrospectively treating the original decision as an assertion of counterfactual truth. Where adjudication identifies a model or knowledge-base deficiency, the finding should inform subsequent model or evidence revision before re-evaluation.

\clearpage
\FloatBarrier
\section*{Supplementary Note 4: Evidence qualification standards}
In Tables S5--S7, Y denotes a core standard within the applicable evidence stream; + denotes the need for strengthened evidence for higher-risk clinical uses (e.g., tighter margins, independent replication or adjudication). A dash indicates no separate enhancement, not exemption from core requirements. These markers are a proposed risk-informed specification, not empirical standards.
\begin{table}[h]
\centering
\caption{Observed-Reality qualification; core standards apply to registered use, with enhanced evidence (+) for higher-risk claims.}
\label{tab:s4observed}
\scriptsize
\begin{tabularx}{\textwidth}{p{2.55cm}X X c c}
\toprule
Standard & Required evidence & Failure consequence & Core & High-risk \\
\midrule
Target-population alignment & Eligibility, setting, calendar period and subgroup coverage match the target population & Restrict claim or perform transport analysis; record unsupported subgroups. & Y & + \\
Well-defined intervention & Treatment version, dose, timing, duration, adherence and switching are encoded & Consistency is not credible. & Y & -- \\
Temporal validity & Confounders precede treatment; branching time and visit order are auditable; no future leakage & Factual and causal checks are invalid. & Y & + \\
Observation-process validity & Censoring, dropout, visit timing, competing events and missing modalities are described and handled relative to the target estimand & Factual visibility may be selectively biased. & Y & + \\
Confounder coverage & Clinically prespecified common causes are measured with adequate quality & Exchangeability remains uncertain; a broader sensitivity analysis is required. & Y & + \\
Sequential support & Required actions have adequate probability over evolving histories and horizons & Unsupported regions are excluded, bounded or Indeterminate. & Y & + \\
Longitudinal depth & Adequate trajectories and events exist at each target horizon & Long-horizon claim is restricted or Indeterminate. & Y & -- \\
Measurement harmonization & Outcomes and mediators are comparable across visits, sites and groups & Distributional comparisons are uninterpretable. & Y & -- \\
Evaluation independence & Patient-level separation and, where possible, external site/period validation & Source-specific fit cannot be excluded. & Y & + \\
Reference causal analysis & Target trial, estimand and estimator are prespecified for causal use & Observed arm contrasts remain descriptive. & Y & + \\
Claim-specific assumptions & Cross-world or pathway assumptions are declared; bounds/sensitivity analyses are prespecified when the target quantity is not identified & personalized or structural claims are overstated. & Y & + \\
Governance and provenance & Data lineage, coding, exclusions and preprocessing are reproducible & Evidence cannot be audited. & Y & + \\
\bottomrule
\end{tabularx}
\end{table}

\begin{table}[h]
\centering
\caption{Qualification standards for Virtual-Reality paired-simulator evidence, when required by the claim. These concern simulator suitability as benchmark evidence, not clinical truth.}
\label{tab:s5virtual}
\scriptsize
\begin{tabularx}{\textwidth}{p{2.55cm}X X c c}
\toprule
Standard & Required evidence & Failure consequence & Core & High-risk \\
\midrule
Structural transparency & State, treatment, outcome, mediator, latent and stochastic mechanisms are documented and versioned & Counterfactual quantities cannot be interpreted. & Y & -- \\
Modular intervention & Replacing the intervention changes only the declared mechanism unless explicitly targeted otherwise & Contrasts are not causally attributable. & Y & -- \\
Paired virtual patients & Baseline and declared latent characteristics are preserved across interventions; noise coupling is specified & Individual/pathwise contrasts are ill-defined. & Y & -- \\
Known counterfactual benchmark & Paired potential outcomes, joint distributions, individual contrasts or pathways required by the claim are exposed & No direct recovery test exists for that claim. & Y & -- \\
Assumption alignment & Confounding, support, coupling and pathway assumptions are explicit and perturbable & Recovery cannot be interpreted against the claim. & Y & + \\
Stochastic replication & Heterogeneity, process noise and repeated patient-intervention trajectories are represented & Uncertainty is artificially narrow. & Y & + \\
Leakage prevention & Exported predictors do not deterministically reveal target outcomes or hidden labels & Recovery may reflect leakage. & Y & + \\
Clinical/temporal admissibility & Generated trajectories respect prespecified ordering and clinical constraints & Benchmark may reward impossible trajectories. & Y & + \\
Sampling transparency & Rejection, truncation, prevalence fixing and balancing are reported & Target population may be distorted. & Y & -- \\
Scenario diversity & Effect size, heterogeneity, confounding, overlap, dynamics and coupling are varied; null controls included & Performance may depend on one favorable simulator. & Y & + \\
Independent calibration & Calibration targets are separated from counterfactual evaluation targets where possible & Circular validation may inflate credibility. & Y & + \\
Reproducibility & Versions, parameters, interventions, seeds and stochastic streams are recorded & Results cannot be reproduced. & Y & -- \\
\bottomrule
\end{tabularx}
\end{table}

\begin{table}[h]
\centering
\caption{Qualification standards for Knowledge-Reality evidence and optional clinician-calibrated medical LLM adjudication (LLM-specific checks apply only when used).}
\label{tab:s6knowledge}
\scriptsize
\begin{tabularx}{\textwidth}{p{2.55cm}X X c c}
\toprule
Standard & Required evidence & Failure consequence & Core & High-risk \\
\midrule
Domain authority & Relevant clinician expertise and/or curated, revisable evidence define admissible knowledge & No defensible reference standard. & Y & + \\
Evidence grounding & Authoritative, versioned sources; when an LLM is used, retrieval-augmented generation over a frozen, cited medical corpus and, where useful, a versioned ontology/knowledge graph, with prespecified retrieval configuration & judgment may rely on unsupported model memory or changing evidence. & Y & -- \\
Task limitation & The clinician-calibrated medical large language model judges clinical, temporal, pharmacological and pathway plausibility; never the missing potential outcome and never as an independent Reality & Epistemic misuse or false evidential independence. & Y & -- \\
Clinician-defined rubric & Hard violations, graded criteria, abstention conditions and escalation rules are prespecified & Subjective/non-reproducible evaluation. & Y & -- \\
Human calibration & Error characteristics are measured on held-out clinician-labeled cases using sensitivity, specificity, abstention, clinically weighted loss and a prespecified false-concordance operating target & Judge reliability and unsafe endorsement risk are unknown. & Y & + \\
Blind evaluation & Generator identity is hidden and presentation order randomized & Preference bias may contaminate results. & Y & + \\
Robustness tests & Position, wording, demographic, prompt and retrieval perturbations are assessed under a prespecified protocol & Judge may respond to irrelevant features or unstable retrieval. & Y & + \\
Adjudication policy & Aggregation, abstention and conflict rules are chosen from held-out clinician-labeled performance and clinical loss; the medical LLM cannot override a valid mandatory contradiction & Arbitrary aggregation or compensatory override. & Y & + \\
Abstention and escalation & Safety-critical flags, insufficient/conflicting retrieval, materially prompt-sensitive judgment, disagreement, abstention or failure to meet the calibrated operating target trigger human review; unresolved review is Indeterminate & Unsafe automated endorsement or false concordance. & Y & -- \\
Version control & Model, prompt, rubric, corpus, retrieval configuration, knowledge graph and decoding settings are frozen and logged; updates trigger review & judgment cannot be reproduced or re-evaluated. & Y & -- \\
Privacy/governance & Identifiable data are restricted to approved systems and evidence use is auditable & Clinical use is not admissible. & Y & + \\
\bottomrule
\end{tabularx}
\end{table}

\clearpage
\FloatBarrier
\section*{Supplementary Note 5: Hypothetical worked example of the CPC decision procedure}
\label{sec:s_worked_example}

\paragraph{Purpose and scope.}
This entirely hypothetical example illustrates how CPC links a registered
counterfactual claim to qualified evidence, mandatory checks and a
non-compensatory decision. It builds on the illustrative registration
in Table~\ref{tab:s3registration} and follows the six stages in
Supplementary Note~3. All numerical inputs, evidence qualifications,
and assessment outcomes below are illustrative stipulations; the
calculations demonstrate their logical consequences. No patient data
were analysed, no simulator was run, and no clinical effect of statins
on dementia or empirical validation of CPC is asserted.

\paragraph{Stage 1: Claim registration.}
Consider an intervention-conditioned human digital twin (IC-HDT)
estimating the population-average difference in five-year dementia
risk under a fixed statin intervention $a_S$ versus comparator $a_0$.
The registered claim is $\CC$-ATE, with target quantity
\[
\psi_c=\E\!\left[Y^{a_S}-Y^{a_0}\right],
\]
where $Y$ indicates dementia occurrence within five years. A negative
value corresponds to a lower population-average risk under $a_S$
than under $a_0$.

The target population $\mathcal P_c$, intervention contrast
$\mathcal G_c$, horizon $\tau_c$, intended history region $\Omega_c$,
context of use $\chi_c$, and evidence and decision specifications follow
Table~\ref{tab:s3registration}. Identification requires consistency
(A1), sequential exchangeability (A2), and positivity (A3), with
transportability (A4) where applicable. Suppose that the applicable
requirements, mandatory checks, thresholds, evidence qualifications,
multiplicity controls and sensitivity ranges were registered and
justified before the hypothetical outcomes were inspected.
These are \emph{stipulated} conditions, not completed assessments.

\begin{table}[h]
\centering
\caption{Filled toy registration for the dementia--statin $\CC$-ATE example. The values illustrate CPC registration fields and do not assert a clinical statin effect on dementia.}
\label{tab:s3registration}
\scriptsize
\begin{tabularx}{\textwidth}{p{1.7cm}p{4.3cm}X}
\toprule
Group & Registration field & Toy specification \\
\midrule
\multirow{7}{*}{\textbf{Claim}}
& Claim class & $\CC$-ATE: Population Effect; population; non-structural; single-point; non-policy. \\
& Target population $\mathcal P_c$ & Registered dementia--statin target population and relevant subgroups. \\
& Intervention contrast $\mathcal G_c$ & Fixed statin intervention $a_S$ versus comparator $a_0$. \\
& Time horizon $\tau_c$ & Five years. \\
& Causal quantity $\psi_c$ & $\E(Y^{a_S}-Y^{a_0})$. \\
& Intended region $\Omega_c$ & Registered clinical histories for which both interventions are to be assessed; later support restrictions require re-registration. \\
& Assumptions and context of use & $\mathcal A_c$: A1--A3; A4 if transported. $\chi_c$: prespecified intended use and model risk. \\
\midrule
\multirow{7}{*}{\textbf{Evidence}}
& Mandatory checks $\mathcal L_c$ & Factual credibility, Observed-Reality compatibility or evidence-gap grade, Knowledge-Reality admissibility, falsification and provenance audit; Virtual-Reality recovery when paired/pathway evidence is invoked. \\
& Factual statistics $\Phi_F^c$ & Prespecified predictive, calibration, distributional, temporal and uncertainty statistics. \\
& Real-world causal constraint & Prespecified estimate, sensitivity region or bounds for the five-year dementia-risk difference. \\
& Sensitivity and uncertainty & Prespecified $\Gamma$ where required and confidence level $1-\alpha_{\mathrm{RW}}$. \\
& Paired-simulator benchmark & Simulator-defined CC-ATE under matched intervention branches, with simulator family and coupling assumptions declared. \\
& Knowledge Reality & Prespecified clinicians and/or qualified guidelines, reviews, pharmacological resources, ontologies or knowledge graphs. \\
& Provenance and dependence & Sources, processing, simulator versions, knowledge sources and shared dependencies recorded for audit. \\
\midrule
\multirow{7}{*}{\textbf{Decision}}
& Discrepancy criteria & Prespecified factual, real-world and simulator discrepancy measures and scaling rules. \\
& Thresholds & $\varepsilon_{F,c}$, $\varepsilon_{\mathrm{RW},c}$ and simulator tolerances; clinical-indifference threshold and plausible sensitivity ranges when required. \\
& Threshold authority and justification & Registered multidisciplinary evaluation team; responsible authority, clinical/statistical rationale, evidence source and plausible range specified before final evaluation. \\
& Optional clinician-calibrated medical LLM interface & If used, frozen RAG corpus/configuration, prespecified rubric, robustness tests, false-concordance operating target and escalation rule. \\
& Threshold calibration & Prespecified operating target using clinically meaningful margins, independent calibration/reference data, reference distributions, benchmark cases or external evidence, as appropriate; no post-evaluation retuning. \\
& Error and multiplicity control & One confirmatory error budget with a prespecified family-wise procedure across applicable stochastic checks and falsifiers. \\
& CPC decision & Apply Eq.~\eqref{eq:s_gate}; report Pass, Fail or reason-coded Indeterminate, with evidence versions and re-evaluation triggers. \\
\bottomrule
\end{tabularx}
\end{table}

This is a population-effect claim, not a realised individual effect.
For the distinct $\CC$-ITE claim about the realised five-year
contrast in one patient, the same population marginals do not establish
cross-world dependence. Without a defensible A5, that claim receives
I-A even if qualified simulator recovery succeeds.
Although a possible simulator benchmark is described in the registration
table, no paired-simulator evidence is invoked in this particular
assessment. Virtual-Reality recovery is therefore not mandatory here;
it would become mandatory if such evidence were invoked, without
thereby establishing clinical counterfactual truth.

\paragraph{Stage 2: Factual credibility.}
Suppose that the registered factual evaluation satisfies both
conditions associated with Eq.~\eqref{eq:s_factual}:
\[
D_F(\CC_c)\leq q_{F,c},
\qquad
U_{F,c}\leq\varepsilon_{F,c}.
\]
Further stipulate that the factual-statistic family, observation-process
handling, uncertainty procedure and relevant evidence qualifications
are adequate. Under these assumptions the factual check is graded P.
No factual discrepancy or confidence bound was calculated for this
example; factual P is \emph{stipulated} and does not establish
counterfactual credibility.

\paragraph{Stage 3: Counterfactual plausibility.}
\noindent\emph{Observed-Reality compatibility.}
Suppose a hypothetical qualified causal analysis yields a 95\%
reference confidence region for the registered population effect:
\[
C_c=[-0.06,-0.04].
\]
This is an assumption-indexed illustrative constraint, not an observed
or clinically established treatment effect. Let the registered
compatibility tolerance be $\varepsilon_{\mathrm{RW},c}=0.01$ and
use absolute distance. The tolerance-enlarged region is
\[
\begin{aligned}
C_c^{+\varepsilon}
 &=\{u:\dist(u,C_c)\leq0.01\}\\
 &=[-0.07,-0.03],
\end{aligned}
\qquad
\operatorname{width}(C_c^{+\varepsilon})=0.04.
\]
For this example, the prespecified competing, decision-relevant
conclusion is a zero or positive risk difference,
$\mathcal B_c=[0,\infty)$. The registered exclusion-based
informativeness condition is satisfied:
\[
C_c^{+\varepsilon}\cap\mathcal B_c=\varnothing.
\]
Suppose the twin estimate is
$\widehat\psi_{\Theta,c}=-0.05$. Its point-estimate distance is
\[
\dist\!\left(\widehat\psi_{\Theta,c},C_c\right)
=0\leq0.01.
\]
These computations establish point-estimate compatibility and
the displayed informativeness condition \emph{only}. They do not
incorporate uncertainty in the twin estimate, establish sequential
exchangeability, or independently qualify the external causal reference.

For the conditional P scenario, additionally \emph{stipulate} that
a registered uncertainty-aware procedure accounting for uncertainty
in both the twin estimate and the reference constraint is satisfied;
that the reference is qualified and sufficiently informative under
the applicable registered criteria; that A1--A4 are defensible
where applicable; and that no material treatment-support or
sensitivity requirement remains unresolved. Only conditional on
these additional stipulations is Observed Reality graded P.

\medskip
\noindent\emph{Knowledge-Reality admissibility.}
Suppose qualified, versioned biomedical evidence has been reviewed
under the registered admissibility and adjudication procedure, with
no qualified mandatory clinical contradiction and no unresolved material
disagreement. Knowledge Reality is then \emph{stipulated} to receive P.
This is completion of a rejection-only admissibility screen, not
positive evidence that the treatment effect is causally correct.

\medskip
\noindent\emph{Virtual-Reality recovery.}
Virtual Reality is marked ``not required'' for this registration
because paired-simulator evidence is not invoked. A possible
simulator benchmark recorded in Table~\ref{tab:s3registration}
does not by itself supply independent clinical causal support.

\paragraph{Stage 4: Prespecified falsification.}
Suppose all applicable registered falsification procedures are
qualified, sufficiently informative against their specified
minimum material violations, and conducted with the registered
multiplicity-adjusted error control where relevant.
Further suppose no qualified contradiction is detected.
The falsification check is \emph{stipulated} to receive P.
No tests or power calculations were performed here; the grade
indicates hypothetical stress-test coverage, not proof of validity.

\paragraph{Stage 5: Uncertainty and evidence-dependence audit.}
Suppose the registered audit of evidence provenance, shared
dependencies, prespecified sensitivity and threshold ranges,
relevant subgroup support and evidence versions is complete.
Further suppose no unresolved material source conflict,
support deficit or status instability remains. The provenance
and uncertainty check is \emph{stipulated} to receive P.
This does not imply independence among evidence streams or
the absence of unrecognised bias.

\paragraph{Stage 6: Non-compensatory CPC decision.}
The stipulated assessment and its limited numerical calculations
are summarised below. The table is intentionally unnumbered to
preserve all existing supplementary table references.

\begin{center}
\begingroup
\scriptsize
\setlength{\tabcolsep}{3pt}
\renewcommand{\arraystretch}{1.15}
\begin{tabularx}{\linewidth}{@{}p{2.45cm}X p{2.1cm}@{}}
\toprule
Check & Basis of the illustrative assessment & Grade \\
\midrule
Factual credibility &
Registered discrepancy, uncertainty and qualification requirements
stipulated satisfied. & P (stipulated) \\
\addlinespace[2pt]
Observed Reality &
Point compatibility and exclusion-based informativeness calculated;
causal qualification and uncertainty-aware compatibility stipulated.
& P (conditional) \\
\addlinespace[2pt]
Knowledge Reality &
Qualified admissibility review stipulated complete, with no
unresolved mandatory contradiction. & P (stipulated) \\
\addlinespace[2pt]
Falsification &
Registered coverage and error control stipulated adequate, with
no qualified contradiction. & P (stipulated) \\
\addlinespace[2pt]
Provenance and uncertainty &
Registered support, dependence and sensitivity audits stipulated
complete and stable. & P (stipulated) \\
\addlinespace[2pt]
Virtual Reality &
No paired-simulator evidence invoked for this population-effect
claim. & Not required \\
\midrule
\textbf{Overall CPC} &
\textbf{Every applicable mandatory check conditionally receives P.}
& \textbf{P (conditional)} \\
\bottomrule
\end{tabularx}
\endgroup
\end{center}

If every stipulated qualification holds, the non-compensatory
rule in Eq.~\eqref{eq:s_gate} gives
\[
\boxed{G_{\CPC}(\CC_c)=\mathrm P}.
\]
This is a \emph{conditional logical verdict}, not an empirically
established Pass. The displayed arithmetic establishes only numerical
relationships between hypothetical quantities; the remaining
mandatory qualifications have been assumed rather than demonstrated.

\paragraph{Contrasting scenario A: Indeterminate due to inadequate informativeness.}
Retain the original reference interval and tolerance, but consider
an \emph{alternative pre-registered specification} additionally
requiring $\operatorname{width}(C_c^{+\varepsilon})\leq0.03$.
Because the enlarged width is $0.04$,
\[
0.04>0.03.
\]
The width-based informativeness criterion fails, even though the
twin estimate remains numerically compatible. Assuming no
qualified mandatory F, Observed Reality is I-E and the overall
verdict is
\[
\boxed{G_{\CPC}(\CC_c)=\mathrm I\quad(\mathrm{I\!-\!E})}.
\]
This illustrates a different registration, not retrospective
tightening of the original criteria after observing the result.

\paragraph{Contrasting scenario B: Indeterminate due to unresolved identification.}
Return to the original registration, without the additional width
restriction. Suppose sequential exchangeability (A2) cannot be
adequately justified. The numerical compatibility calculation is
unchanged, but causal interpretation of the reference constraint
is insufficiently supported. Observed Reality receives I-A;
provided no mandatory check is F,
\[
\boxed{G_{\CPC}(\CC_c)=\mathrm I\quad(\mathrm{I\!-\!A})}.
\]
Numerical agreement cannot compensate for an unresolved
causal-identification assumption.

\paragraph{Contrasting scenario C: Fail due to a qualified contradiction.}
Consider a different hypothetical twin estimating
$\widehat\psi_{\Theta,c}=+0.06$, while retaining
$C_c=[-0.06,-0.04]$ and tolerance $0.01$:
\[
\begin{aligned}
\dist(0.06,C_c)
 &=0.06-(-0.04)\\
 &=0.10>0.01.
\end{aligned}
\]
The point-estimate compatibility inequality fails. This alone
does \emph{not} establish a qualified F, because uncertainty
in the twin and reference estimates and the credibility
of the causal reference must also be evaluated.

For this scenario, additionally stipulate that the registered
uncertainty-aware assessment confirms incompatibility and
that adjudication establishes a qualified contradiction to
the registered mandatory Observed-Reality criterion.
Under those conditions Observed Reality is F and
\[
\boxed{G_{\CPC}(\CC_c)=\mathrm F}.
\]
Other satisfactory checks cannot compensate for a qualified
mandatory contradiction.

\paragraph{Interpretation and limitations.}
These scenarios illustrate conditional P when every mandatory
requirement is met, I-E when evidence is insufficiently informative,
I-A when causal assumptions remain unresolved, and F following
a qualified mandatory contradiction. The decision rules are
non-compensatory and claim-specific; their outputs remain
conditional on the registered evidence state and assumptions.

The worked example does not estimate a clinical treatment effect,
establish the validity of CPC, calibrate decision thresholds,
measure false-Pass or false-Fail rates, or demonstrate clinical
utility. Only the arithmetic is explicitly worked; evidence
qualification, statistical uncertainty assessment, falsification
coverage and provenance stability are stipulated.
\FloatBarrier
\FloatBarrier
\printbibliography